\pdfoutput=1

\documentclass[11pt]{article}

\usepackage[final]{acl}

\usepackage{times}
\usepackage{latexsym}
\usepackage{subcaption}
\usepackage{float}
\usepackage[T1]{fontenc}

\usepackage[utf8]{inputenc}

\usepackage{microtype}
\usepackage{pifont}
\usepackage{epsfig}
\usepackage{graphicx}
\usepackage{amsmath}
\usepackage{amssymb}
\usepackage{colortbl}
\usepackage{multirow}
\usepackage{diagbox}
\usepackage{color,xcolor}
\usepackage{adjustbox}
\usepackage{makecell}
\usepackage{enumitem}

\usepackage{subcaption,booktabs}
\usepackage{wrapfig}
\usepackage{inconsolata}
\makeatletter
\newcommand{\thickhline}{%
    \noalign {\ifnum 0=`}\fi \hrule height 1pt
    \futurelet \reserved@a \@xhline
}

\usepackage{graphicx}
\definecolor{fired}{RGB}{222,82,57}
\definecolor{fourier_green}{RGB}{0, 128, 0}
\definecolor{visual_yellow}{RGB}{191, 144, 0}
\definecolor{visual_prompt_yellow}{RGB}{246, 212, 109}
\definecolor{text_prompt_pink}{RGB}{250, 204, 228}
\definecolor{iceblue}{RGB}{33,102,200}
\definecolor{class}{HTML}{bb9726}
\definecolor{prompt}{HTML}{2978b5}
\definecolor{input}{HTML}{53b245}
\definecolor{mygray}{gray}{.9}
\definecolor{fired}{RGB}{222,82,57}
\definecolor{iceblue}{RGB}{33,102,200}

\newcommand{\ie}{\textit{i}.\textit{e}.}
\newcommand{\eg}{\textit{e}.\textit{g}.}

\title{M$^2$PT: Multimodal Prompt Tuning for Zero-shot Instruction Learning}

\author{
 \textbf{Taowen Wang\textsuperscript{1}\thanks{Equal contribution.}},
 \textbf{Yiyang Liu\textsuperscript{1}\footnotemark[1]},
 \textbf{James Chenhao Liang\textsuperscript{1,11}},
 \textbf{Junhan Zhao\textsuperscript{2}},
\\
 \textbf{Yiming Cui\textsuperscript{3}},
 \textbf{Yuning Mao\textsuperscript{4}},
 \textbf{Shaoliang Nie\textsuperscript{4}},
 \textbf{Jiahao Liu \textsuperscript{5}},
\\
 \textbf{Fuli Feng\textsuperscript{6}},
 \textbf{Zenglin Xu\textsuperscript{7,8}},
 \textbf{Cheng Han\textsuperscript{9}},
 \textbf{Lifu Huang\textsuperscript{10}},
\\
 \textbf{Qifan Wang\textsuperscript{4}},
 \textbf{Dongfang Liu\textsuperscript{1}}
\\
\\
 \textsuperscript{1}Rochester Institute of Technology,
 \textsuperscript{2}Harvard Medical School,
 \textsuperscript{3}ByteDance,
 \textsuperscript{4}Meta AI,
 \textsuperscript{5}Meituan,
\\
 \textsuperscript{6}University of Science and Technology of China,
 \textsuperscript{7}Shanghai Academy of AI for Science,
\\
 \textsuperscript{8}Fudan University,
 \textsuperscript{9}University of Missouri - Kansas City,
 \textsuperscript{10}University of California - Davis
\\
 \textsuperscript{11}U.S. Naval Research Laboratory
\\
 \small{
    \{tw9146, dongfang.liu\}@rit.edu}
 }

\begin{document}
\maketitle
\begin{abstract}

Multimodal Large Language Models (MLLMs) demonstrate remarkable performance across a wide range of domains, with increasing emphasis on enhancing their zero-shot generalization capabilities for unseen tasks across various modalities. Instruction tuning has emerged as an effective strategy for achieving zero-shot generalization by finetuning pretrained models on diverse multimodal tasks. As the scale of MLLMs continues to grow, parameter-efficient finetuning becomes increasingly critical. However, most existing parameter-efficient approaches focus only on single modalities and often overlook the multimodal characteristics during finetuning. In this work, we introduce a novel Multimodal Prompt Tuning (M$^2$PT) approach for efficient instruction tuning of MLLMs. M$^2$PT effectively integrates visual and textual prompts into the vision encoder and language processor respectively during finetuning, facilitating the extraction and alignment of features across modalities. Empirical results on various multimodal evaluation datasets demonstrate the superior performance of our approach compared to several state-of-the-art baselines. A comprehensive set of ablation studies validates the effectiveness of our prompt design and the efficiency of our approach.
\end{abstract}

\section{Introduction}\label{sec:intro}
Human cognition intricately integrates various sensory modalities to perceive, interpret, and engage with the environment, fostering a comprehensive understanding of the surrounding world~\cite{liu2024visual,dumas2009multimodal,Chen2024Inertial,jin2024prollm,Jin2024Health}. 
The development of Multimodal Large Language Models (MLLMs)~\cite{alayrac2022flamingo,yin2023survey,Han2024Image,Han2024AMD,Jin2024Exploring,Jin2024impact} marks a significant advancement in emulating this capability, playing a pivotal role in bridging the gap between human and machine intelligence. A key focus in advancing MLLMs is enhancing their zero-shot generalization to new multimodal tasks. In this pursuit, multimodal instruction tuning~\cite{liu2024visual,abs-2310-03744,XuSH23}, which finetunes pretrained models using diverse and instruction-based multimodal tasks, has proven effective in improving zero-shot generalization to unseen multimodal domains.

\begin{figure}[t!]
    \centering
    \includegraphics[width=0.45\textwidth]{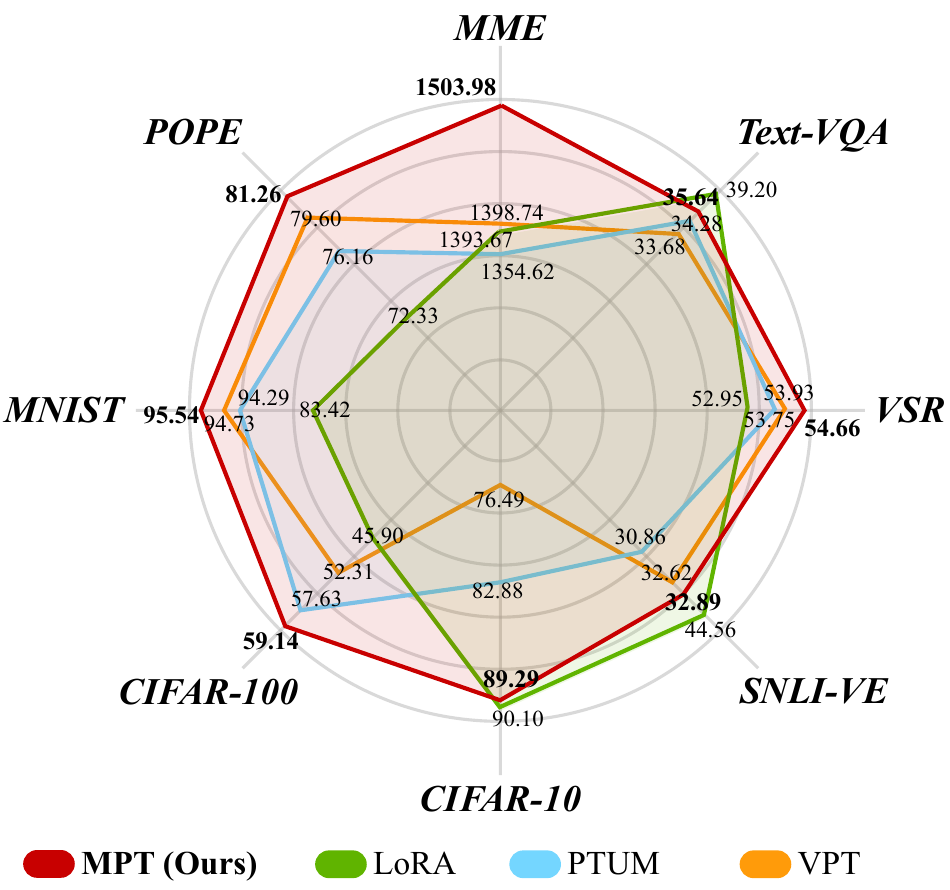}
    \caption{\textbf{Comparison of M$^2$PT and several PEFT methods}, including LoRA~\cite{hu2021lora}, PTUM~\cite{PTUMPM} and VPT~\cite{han2024facing}, on multimodal tasks. Our approach exhibits superior performance across a range of benchmarks. }
    \vspace{-4mm}
    \label{fig:question_exp}
\end{figure}

Despite considerable advancements, finetuning MLLMs for specific domain knowledge poses significant challenges. As the scale and complexity of these models increase, the training overhead for downstream tasks grows exponentially~\cite{DBLP:journals/corr/abs-2403-15226, DBLP:journals/corr/abs-2404-16994,zhang2024stock}. These escalating demands render the current tuning schemes for MLLMs obsolete and unsustainable, impeding their widespread utility. A promising solution is the utilization of parameter-efficient finetuning (PEFT) approaches~\cite{lester2021power,hu2021lora,zang2022unified,dong2024efficient,aprompt}, which have been widely applied and achieved notable success in natural language processing~\cite{liu2021p, DBLP:conf/eccv/0002ZESZLRSPDP22} and computer vision tasks~\cite{jia2022visual, han20232vpt, DBLP:conf/eccv/JuHZZX22}. However, most existing PEFT approaches only focus on single modality tuning while overlooking multimodal instructing learning. A \textit{primary challenge} is the intricate process of finetuning data of multiple modalities within a cohesive model, extracting and aligning feature representations across modalities. Additionally, there is a pressing need to enhance the zero-shot generalization capabilities for unseen multimodal tasks while minimizing training costs.

In light of this, we present a novel \textbf{M}ulti\textbf{m}odal \textbf{P}rompt \textbf{T}uning (M$^2$PT) approach with efficient and effective MLLM adaptation for zero-shot instruction learning.
M$^2$PT demonstrates competitive performance across a wide spectrum of tasks (see Fig.~\ref{fig:question_exp}) while tuning 0.09\% of overall parameters.
Specifically, we introduce two sets of soft prompts: visual prompts and textual prompts, which are prepended to the visual and instruction inputs, respectively. The learned embeddings of the visual prompts are projected into the embedding space of the textual prompts. The cross-modality interactions between the two sets of prompts are enforced during instruction tuning, facilitating the alignment and learning of the feature representation between the two modalities. In this way,
M$^2$PT provides explicit guidance and clear directives through instruction tuning, enabling models to understand context and reduce ambiguity in zero-shot settings.

To effectively assess our method, we conduct comprehensive experiments to evaluate its performance. In \S\ref{subsec:main-results}, we demonstrate that M$^2$PT surpasses several state-of-the-art PEFT techniques while tuning only 0.09\% of the trainable parameters. We further conduct activation analysis to show the effectiveness of the learned prompts during the attention computation. In \S\ref{subsec:ablation}, we provide various ablation analysis on the impact of model components, prompt length, prompt location, and data volumes in detail. Furthermore, we conduct case studies to better understand the advantage of M$^2$PT and its limitations.
We hope this work offers valuable insights into related fields. We summarize
the main contributions as follows:
\begin{itemize}
\item We present multimodal prompt tuning by introducing both visual and textual prompts into vision encoder and language processor respectively. These modality specific prompts play a crucial role in effectively guiding the model's fine-tuning process, enabling fast and accurate multimodal model adaptation.
\item We design the cross-modality interaction between the prompts from different modalities during the instruction tuning. By doing so, M$^2$PT leverages the strengths of each modality, resulting in comprehensive and coherent learning results. This synergy empowers the model to perform complex tasks that require the integration from multimodal perspectives.
\item We conduct comprehensive experiments on various multimodal tasks in the zero-shot setting, demonstrating the effectiveness of the proposed approach over several state-of-the-art parameter efficient finetuning methods.
\end{itemize}

\section{Related Work}\label{sec:related_work}

\paragraph{Multimodal Large Language Models.} 

MLLMs \cite{dai2024instructblip, driess2023palme, liu2024visual,YaoZX23,sun2024visualagentsfastslow} integrate multimodal information (\eg, audio, image, video),
extending beyond the textual semantic information processed by conventional Large Language Models (LLMs).A general structure of MLLMs includes three main components~\cite{li2024multimodal}: a pre-trained modality encoder to encode multimodality data, a pre-trained LLM to reason encoded multimodal data and perform generation tasks, and a connection layer to project modality information into tokens. During the standard full finetuning process~\cite{liu2024visual}, a substantial amount of weights within all intermediate layers and the pre-trained LLM are continuously updated. Conceptually different, our approach can elegantly fine-tune the model with adjusting a minimum of weights.

\paragraph{Instruction Tuning.} 
To enhance the zero-shot and in-context learning~\cite{brown2020language,li2024finetuning} capabilities of large language models (LLMs)~\cite{zhao2023survey}, researchers have explored instruction tuning~\cite{ouyang2022training,zhang2023instruction}, a technique that enables pre-trained LLMs to be more adaptable for intricate multimodal tasks. Specifically, instruction-tuning is a process that refines LLMs by finetuning them on meticulously curated instruction-following datasets encapsulating user intent and desired outputs~\cite{ouyang2022training}. With the rapid advancement of multimodal models,  
instruction tuning has emerged not only as a state-of-the-art approach for natural language processing (NLP) tasks but also naturally extend to vision-related tasks such as image captioning~\cite{image-cap, image-cap2}, image classification~\cite{clip2021}, visual question answering (VQA)~\cite{antol2015vqa}.

\paragraph{Parameter-Efficient Finetuning.} 
With the drastic growth in the scale of AI models, especially MLLMs and LLMs, Parameter-efficient Finetuning (PEFT)~\cite{hu2021lora, he2023parameter, jie2023revisiting, Yan24Prompt} have shown its capability to efficiently adapt pre-trained models to diverse downstream tasks without updating significant amount of model parameters. Generally, current PEFT strategies can be categorized into \textit{partial tuning}~\cite{chen2021empirical,jia2021exploring}, \textit{extra module} such as Low-Rank Adaptation (LoRA)~\cite{jie2023revisiting} and \textit{prompt tuning}~\cite{jia2022visual, ju2022prompting, dong2023efficient}. However, partial tuning faces limitations, including unsatisfactory performance relative to full finetuning~\cite{jia2022visual, chen2021empirical,jia2021exploring}. Extra module exhibits limited adaptability when considering various model backbones. In contrast, prompt tuning~\cite{lester2021power, ma2022xprompt, he2022hyperprompt,liu2023pre, qiu2020pre} provides a general and straightforward solution with powerful performance gains. It introduces a set of learnable parameters to the input sequence of backbone models, updating only these parameters during finetuning. Despite its simplicity, the applicability of prompt tuning within the multimodal paradigm remains largely unexplored.
Unlike approaches such as MaPLe~\cite{MaPLe}, CoOp~\cite{zhou2022coop}, CoCoOp~\cite{zhou2022cocoop} and MoPE~\cite{Wu2024MoPE}, which focus on crafting CLIP-based prompts for classification tasks, our work targets enhancing the capabilities of MLLMs in zero-shot instruction-following scenarios, resulting in a fundamentally distinct method design. 
Furthermore, PromptFuse~\cite{liang2022PromptFuse} only introduces learnable prompts into textual modality, neglecting the synergy of multimodality. Our approach offers flexibility in prompt design, allowing prompts to be independently tailored for each modality and inserted at various layers. This flexibility significantly enhances the MLLMs' adaptability while reducing the number of parameters, improving performance across various datasets.

\begin{figure*}[!tbh]
    \centering
    \includegraphics[width=0.98\textwidth]{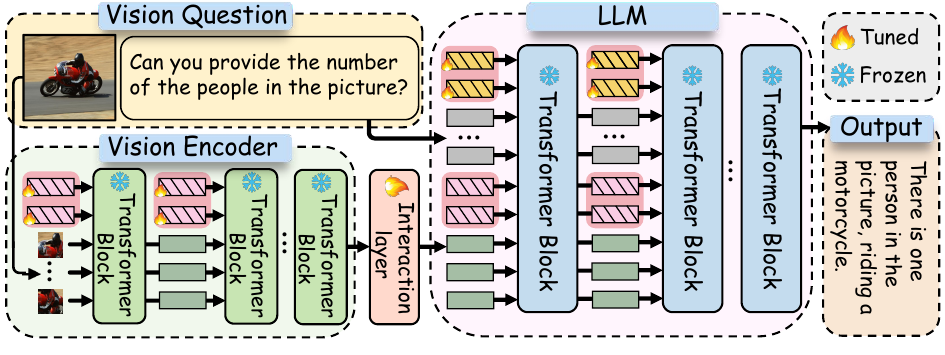}
    \vspace{-2mm}
    \caption{
    \textbf{Overview of our M$^2$PT approach.} 
    Here, \textcolor{text_prompt_pink}{visual} prompts are embedded into each layer of the \emph{Visual Encoder}, and \textcolor{visual_prompt_yellow}{textual} prompts are embedded into each layer of the \emph{LLM}. These prompts facilitate the extraction and alignment of features across modalities (\eg, vision, language). The cross-modality interaction between visual and textual features is enhanced through layered integration, ultimately improving the model's capability in \emph{zero-shot} instruction learning tasks (see \S\ref{sec:exp}).
    }
    \vspace{-4mm}
    \label{fig:mainfig}
\end{figure*}
\section{Methodology}\label{sec:method}

\subsection{Preliminaries}
\paragraph{Multimodal Large Language Models} integrate visual and language processing capabilities, leveraging the power of LLMs to enhance the comprehension of multimodal information. A prevalent workflow of MLLMs begins with the utilization of pre-trained vision encoders $f_{v}$ (\eg, LLaVA~\cite{liu2024visual} and its variants~\cite{blip2022Junnan,ecalclip2023}), encoding visual input $X_{im}$ and extracting output $O_v = f_v(X_{im})$ through remarkable vision understanding ability. Subsequently, the vision output is further projected into language space, aligning with the textual embedding and enabling the model to understand and respond to instructions effectively. Ultimately, the integrated LLM $f_{llm}$ assimilates $O_v$ and text embedding $O_t$. Harnessing the extensive knowledge of LLM, integrating multimodal inputs to generate coherent and contextually relevant language response $y$, represented as:
\begin{equation}
    y = f_{llm}(O_v,O_t).
\label{eq:llm_forward}
\end{equation}

\noindent \textbf{Prompt Tuning} is a form of PEFT approach, demonstrating exceptional efficacy within single-modality settings under both visual~\cite{han20232vpt} and textual~\cite{aprompt} domains. It entails learnable continuous soft prompts into the input space while concurrently preserving the majority of parameters within the backbone frozen.
Specifically, given a $K$ layer transformer-based model $f$, soft prompts $P^k$ combined with the input of $k$-th layer to obtain the output $O^{k}$ as:
\begin{equation}
\begin{aligned}
    &O^1 = \textcolor{iceblue}{f^1}(\textcolor{fired}{P^1}, \ \textcolor{iceblue}{E}) \\
    &O^k = \textcolor{iceblue}{f^k}(\textcolor{fired}{P^k}, \ O^{k-1}),
\end{aligned}
\end{equation}
where $k \in \{2, 3,\dots,K\}$. \textcolor{iceblue}{$\bullet$} and \textcolor{fired}{$\bullet$} indicate frozen and tunable parameter during finetuning, respectively. $E$ is the embedded vector of initial inputs.

\subsection{Multimodal Prompt Tuning}
\label{subsec:MMPT}
In this section, we formally introduce M$^2$PT, a novel multimodal prompt tuning approach for the effective and efficient finetuning of MLLMs. The overall model architecture is depicted in Figure \ref{fig:mainfig}. Fundamentally, our model necessitates the training of only three targeted components, while keeping the backbone parameters from both visual encoder and LLM frozen. Specifically, these components include \ding{172} \textit{Visual Prompt} ($P_v$), which is the learnable parameter (\ie, soft prompt) integrated into the visual encoder; \ding{173} \textit{Textual Prompt} ($P_t$) is incorporated into the LLM in order to capture the nuanced semantic relationships across modalities; \ding{174} \textit{Cross-modality Interaction}, where parameters are learned to enhance the alignment between features extracted by the vision encoder and textual representations. In summary, prompts from these distinct modalities (\ie, \ding{172}-\ding{173}) facilitate the model's acquisition of knowledge from multimodal finetuning datasets, capturing the distinctive characteristics inherent in each modality and fostering cross-modal interaction.

\noindent \textbf{Visual Prompt.} We denote the set of visual prompts as $P_v=\{P_v^1,P_v^i,\cdots,P_v^N\}$, where $P_v^i$ indicates the set of visual prompts in the $i$-th layer of the visual encoder, consistent with previous practice for prompt tuning \cite{jia2022visual, han20232vpt}. Each prompt is a $d_v$-dimensional vector with the same dimensions as the original vision tokens. Prompts in each layer are placed at the leftmost positions within the sequence to interact with other vision tokens.  Formally, we have:

\begin{equation}
\begin{aligned}
    &O_v^1 = \textcolor{iceblue}{f_v^1}(\textcolor{fired}{P_v^1}, \ X_{im}) \\
    &O_v^i = \textcolor{iceblue}{f_v^i}(\textcolor{fired}{P_v^i}, \ O^{i-1}), \\
\end{aligned}
\end{equation}
where $i \in \{2, 3,\dots,N\}$, and $O_v^i$ is the $i$-th layer vision embedding.

\noindent \textbf{Textual Prompt.} Visual prompts serve as an effective tool for capturing semantic content within the visual inputs, as well as gaining the ability to interact with the text modality through the mapping from visual domain. Nevertheless, the optimization of visual elements does not directly affect the inherent representation of LLMs in text modality. Naturally, to further enhance the text modality's processing ability, we introduce the textual prompts to capture text patterns and influence the inner representation within the LLM. Specifically, textual prompts are denoted as $P_t=\{P_t^1,P_t^j,\cdots,P_t^M\}$, where $j$ indicates the prompt inject position of the $j$-th layer in a total $M$ layers LLM. Each prompt is a $d_t$-dimensional vector with the same dimensionality as the original text tokens. Formally, we incorporate textual prompts into the LLM as:

\begin{equation}
\begin{aligned}
    O_m^1 &= \textcolor{iceblue}{f_{llm}^1}(\textcolor{fired}{P_t^1}, \ O_v^{'}, \ \textcolor{iceblue}{O_t}) \\
    O_m^j &= \textcolor{iceblue}{f_{llm}^j}(\textcolor{fired}{P_t^j},\ O_m^{j-1}) \\
    y &=\textcolor{fired}{f_{head}}(O_m^j),
\end{aligned}
\end{equation}
where $j \in \{2, 3,\dots,M\}$, $y$ is the textual output of the MLLM, $O_t$ is the textual embedding, and $f_{head}$ is the task-specific head in order to decode the embeddings into texts.

\noindent \textbf{Cross-modality Interaction.} To achieve alignment between visual and textual modality, we introduce a tunable interaction layer $f_{in}$, which is specifically designed to align the output $O_v^N$ produced by the visual encoder and the textual embedding, through a linear projection:
\begin{equation}
\begin{aligned}
    &O_v^{'} = \textcolor{fired}{f_{in}}(O_v^N). \\
\end{aligned}
\end{equation}

\begin{table*}[t]
\caption{\textbf{Zero-shot Multimodal Evaluation on all multimodal datasets.} The MMAvg represents the average score on the right seven tasks. LLaVA$_{Align}$ is the stage-one LLaVA without end-to-end finetuning, and  LLaVA$_{FT}$ indicates the fully fine-tuned LLaVA. All the fine-tuned processes are using the same Vision-Flan dataset. M$^2$PT$_{a/b}$ means textual and visual prompt lengths `a' and `b', respectively. The best performance is in \textbf{bold}.}
\label{table:main-results}
\vspace{-1mm}
\begin{adjustbox}{width=0.7\width,center}
\begin{tabular}{c|c|c|ccccccccc} 
\toprule
\rowcolor{mygray}
Method & \# para & MME&Text-VQA&VSR&SNLI-VE&CIFAR-10 & CIFAR-100 & MNIST & POPE & MMAvg \\
\midrule
LLaVA$_{Align}$~\cite{liu2024visual} & - &
\cellcolor{mygray} 1110.82 & 32.62 & 50.16 & 34.51 & 80.00 & 58.04 & 52.79 & 59.10 & \cellcolor{mygray} 52.46\\
LLaVA$_{FT}$~\cite{liu2024visual} & 100\% & \cellcolor{mygray} 1587.26 & 37.26 & 53.76 & 43.35 & 92.97 & 63.73 & 94.27 & 80.82 & \cellcolor{mygray} 66.59 \\
\midrule
LoRA~\cite{hu2021lora} & 0.63\%  &  \cellcolor{mygray} 1393.67 & \textbf{39.20} & 52.95 & \underline{44.56} & \textbf{90.10} & 45.90 & 83.42 & 72.33 & \cellcolor{mygray} 61.21\\

APrompt~\cite{aprompt} & 0.23\% & \cellcolor{mygray} 1406.63  & 35.26 & 53.12   & \textbf{45.58}  &  85.74 & 50.27 & 84.63 & 76.16 & \cellcolor{mygray} 61.52 \\
PTUM~\cite{PTUMPM} & 0.12\% & \cellcolor{mygray} 1354.62  & 34.28 & 53.75   & 30.86  &  82.88 & 57.63 & 94.29 & 80.31 & \cellcolor{mygray} 62.00 \\
VPT~\cite{han2024facing} & 0.06\% & \cellcolor{mygray} 1398.74 &  33.68 & \underline{53.93}   & 32.62  &  76.49 & 52.31 & \underline{94.73} & 79.60 &  \cellcolor{mygray} 60.48 \\
M$^2$PT$_{10/10}$ (Ours) & 0.08\% & \cellcolor{mygray} \underline{1490.17} & \underline{35.64} & \textbf{54.66}   & 32.53  &  87.92 & \underline{57.80}  & 94.53 & \textbf{81.29} & \cellcolor{mygray} \underline{63.48} \\
M$^2$PT$_{10/20}$ (Ours) & 0.09\% & \cellcolor{mygray} \textbf{1503.98} & 34.48 & 53.19   & 32.89  &  \underline{89.29} & \textbf{59.14} & \textbf{95.54} & \underline{81.26} & \cellcolor{mygray} \textbf{63.68} \\
\bottomrule
\end{tabular}
\end{adjustbox}
\vspace{-3mm}
\end{table*}

Here $O_v^{'}$ represents the aligned vision embedding. This transformation ensures that the visual encoder's output is effectively mapped onto a common textual representation space. The projected visual tokens then interact with the textual tokens through all LLM layers, facilitating the integration of visual and textual information. Our elegant design of M$^2$PT enjoys a few appealing characteristics:

\begin{itemize}[leftmargin=*]
    \item \textit{Cross-modal Integration.} Our M$^2$PT model employs a unified prompt tuning paradigm. This approach not only captures the distinctive attributes of each modality but also facilitates the fluid exchange of cross-modal information, enhancing the model's capability to comprehend and generate multimodal data effectively.
    \item \textit{Optimized Parameter Utilization.}  M$^2$PT demonstrates superior parameter efficiency by focusing \textit{only} on the training of a minimal set of parameters while keeping the majority of the model's parameters frozen, allowing a significant reduction in the number of parameters required (0.09\%). Despite this reduction, M$^2$PT maintains superior performance on multimodal tasks (see Table~\ref{table:main-results}) with a balance between computational efficiency and overall effectiveness in zero-shot setting.
\end{itemize}

\subsection{Implementation Details}\label{sec:arch}
We employ LLaVA~\cite{liu2024visual} with CLIP-L~\cite{clip2021} (\ie, 24 transformer blocks) as the visual encoder and
Vicuna-7B-v1.3~\cite{zheng2024judging} (\ie, 32 transformer blocks) as the base LLM for all variants.
For the cross-modality interaction, we use a linear layer to map the embedding dimension from $d_v$ to $d_t$ to ensure the alignment between the two modalities. For the prompt initialization, we employ Xavier~\cite{xavier} initialization on both visual and textual prompt to ensure stable modal information delivery from these prompts at the early stages of training, thereby facilitating rapid convergence of the model. More implementation details are provided in \S\ref{subsec:exp-setup} and Appendix \ref{Appendix:Training-detail}.

\section{Experiments}\label{sec:exp}

\subsection{Experiment Setup}\label{subsec:exp-setup}
\paragraph{Datasets.} For \textit{training}, we conduct instruction tuning on \textit{Vision-Flan}~\cite{xu2024vision}, which is a human-annotated multimodal instruction tuning dataset with $191$ diverse tasks. To reduce computational costs, we follow common practice~\cite{shen2024multimodal} and employ a scaled-down version containing up to $1,000$ instances per task, resulting in a total of $191,105$ instances. For \textit{zero-shot evaluation}, we examine our approach on the comprehensive multimodal evaluation benchmark \textit{MME}~\cite{Fu2023MME}, measuring both perception and cognition abilities across $14$ subtasks. We further evaluate the model’s capabilities using $7$ multimodal datasets. Specifically, for Optical Character Recognition, we utilize the \textit{Text-VQA}~\cite{singh2019towards}, and for reasoning, we employ the Visual Spatial Reasoning (\textit{VSR})~\cite{liu2023visual}. Following~\cite{zhai2023investigating, shen2024multimodal}, the perception capability is tested on \textit{CIFAR-10/100}~\cite{krizhevsky2009learning} and \textit{MNIST}~\cite{deng2012mnist}. \textit{SNLI-VE}~\cite{xie2019visual} evaluates Visual Entailment capabilities, while the \textit{POPE}~\cite{li2023evaluating} dataset examines the tendency towards object hallucination. More details are provided in the Appendix \ref{appendix:data-detailed}. 

\noindent \textbf{Training Details.} 
Following previous works~\cite{han20232vpt, shen2024multimodal, jia2022visual}, 
we conduct grid search to match the best tuning hyperparameters, learning rate (\ie, [$1e^{-3}$, $9e^{-4}$, $7e^{-4}$, $4e^{-4}$, $1e^{-4}$, $5e^{-5}$]), textual prompt length (\ie, [0, 5, 10, 20, 40]) and visual prompt length (\ie, [0, 5, 10, 20, 40]). For all models, the learning rate is scheduled following a cosine decay policy, the warm up ratio is set at 0.03 and trained for 3 epochs except in training epoch experiment. We follow the same batch size setting in \cite{shen2024multimodal} as 128 for instruction tuning. Further details are provided in the Appendix \ref{appendix:training}. M$^2$PT is implemented in Pytorch~\cite{NEURIPS2019_9015}. 
Experiments are conducted on 8 NVIDIA A100 GPUs. Our code is available at \url{https://github.com/William-wAng618/M2PT}.

\noindent \textbf{Evaluation Metrics.}
The MME incorporates both Perception and Cognition metrics\footnote{https://github.com/BradyFU/Awesome-Multimodal-Large-Language-Models/tree/Evaluation}. For other multimodal datasets, we use Vicuna-13B-v1.5~\cite{zheng2024judging} to assess the accuracy of each prediction compared to the groundtruth. Further details are provided in the Appendix \ref{appendix:eval}.

\begin{figure}[t]
\vspace{-10pt}
    \centering 
    \begin{minipage}[b]{0.23\textwidth}
        \includegraphics[width=\textwidth]{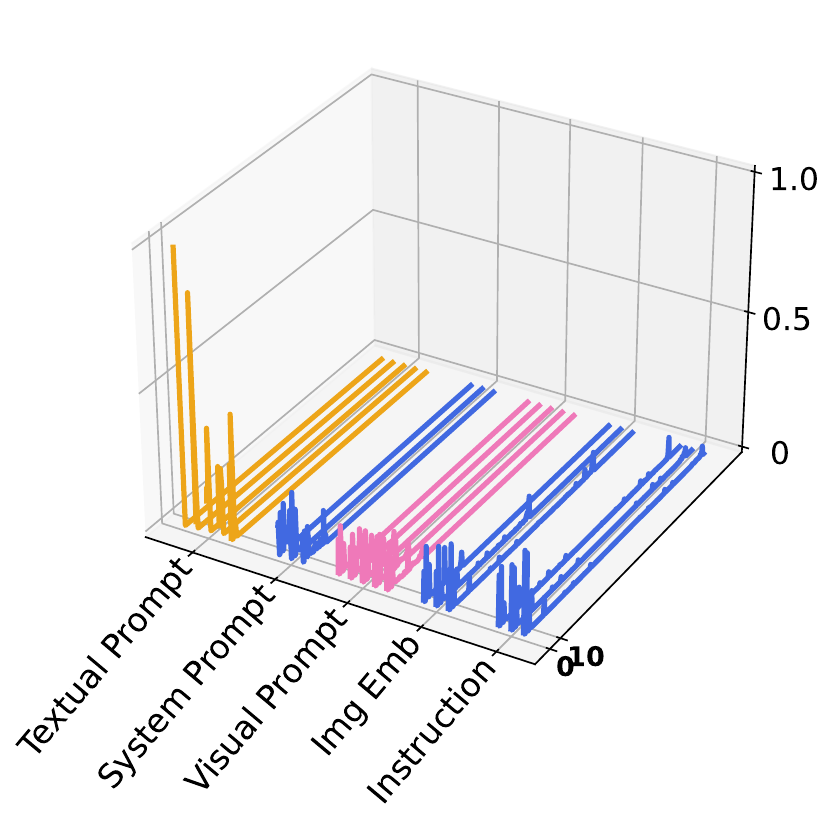}
        \label{fig:image1}
    \end{minipage}
    \begin{minipage}[b]{0.23\textwidth}
        \includegraphics[width=\textwidth]{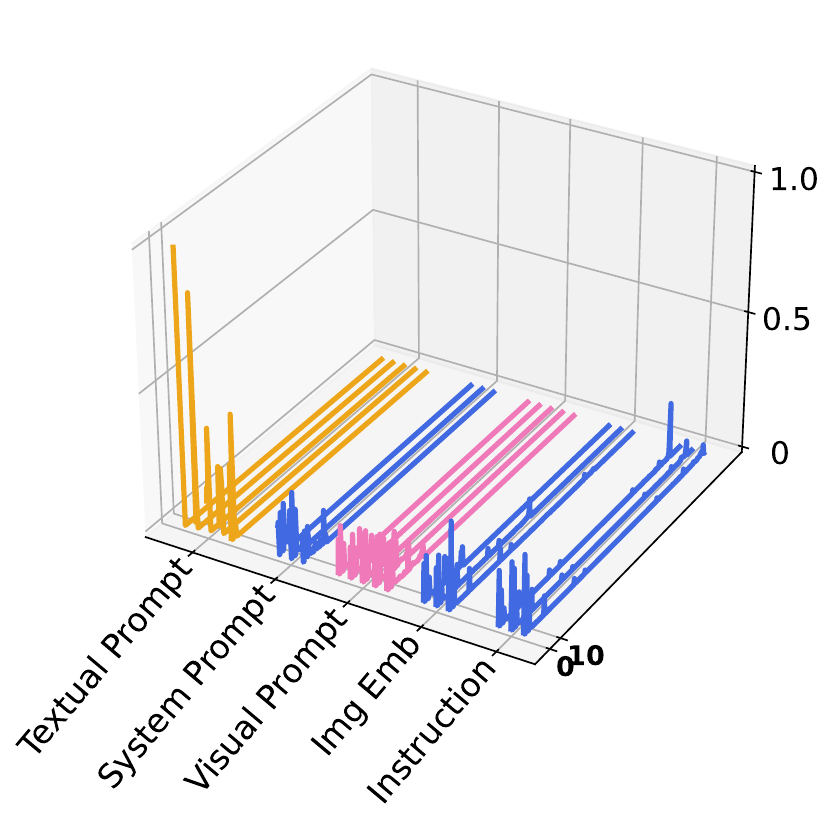}
        \label{fig:image2}
    \end{minipage}
    \vspace{-17pt}
    \captionof*{figure}{(a) \textit{LLM} attention activation}

    \begin{minipage}[b]{0.23\textwidth}
        \includegraphics[width=\textwidth]{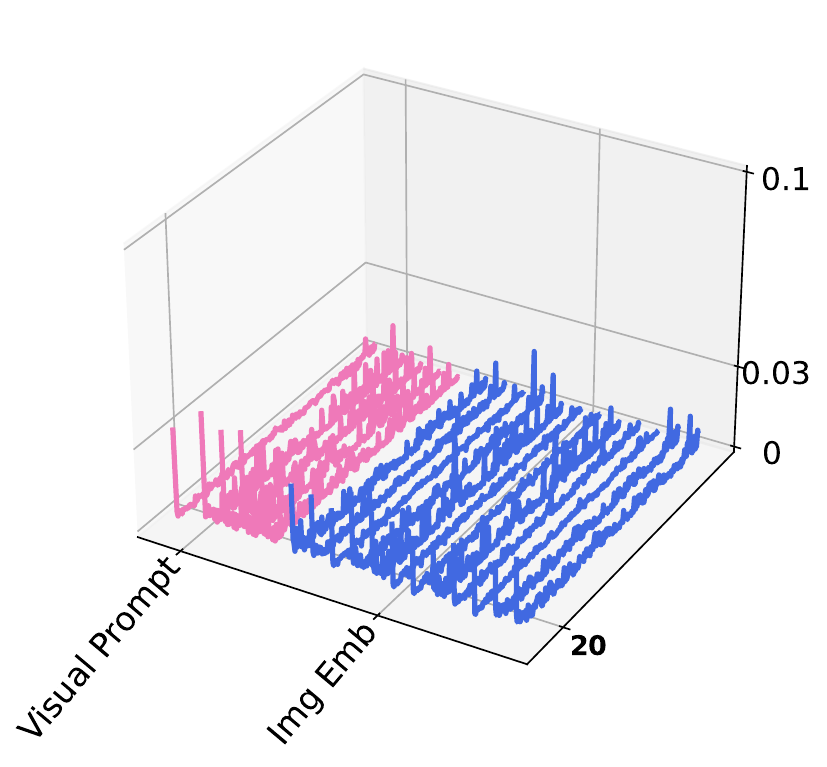}
        \label{fig:image3}
    \end{minipage}
    \begin{minipage}[b]{0.23\textwidth}
        \includegraphics[width=\textwidth]{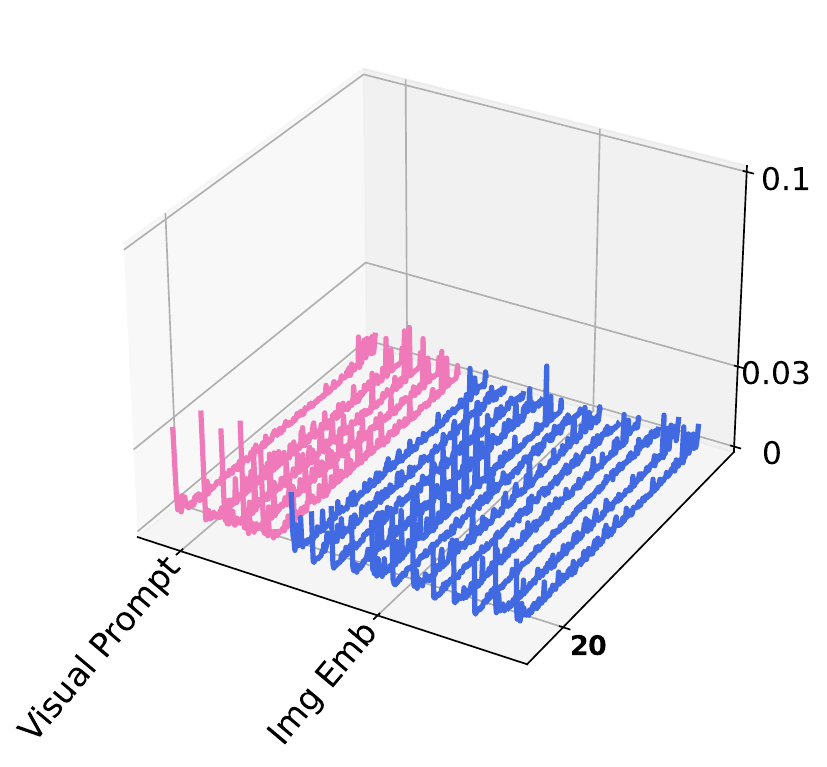}
        \label{fig:image4}
    \end{minipage}
    \vspace{-17pt}
    \captionof*{figure}{(b) \textit{Visual Encoder} attention activation}
    \vspace{-5pt}
    \caption{\textbf{Comprehensive visualization of attention activation maps}. This figure presents a detailed examination of the activation patterns within the last layer of \textit{LLM} and \textit{Visual Encoder}, respectively. As seen, the vision prompts and textual prompts have noticeably high activation levels during inference (\ie, \textcolor{visual_prompt_yellow}{$\bullet$} and \textcolor{text_prompt_pink}{$\bullet$} represent textual prompts' activation signal and visual prompts' activation signal, respectively). }
    \label{fig:attention}
    \vspace{-1.5em}
\end{figure}
\subsection{Main Result}\label{subsec:main-results}
In Table~\ref{table:main-results}, our main result exhibits a comprehensive zero-shot evaluation of M$^2$PT with several baselines on eight multimodal datasets. 
Specifically, we consider four state-of-the-art PEFT approaches, including LoRA~\cite{hu2021lora}, APrompt~\cite{aprompt}, PTUM~\cite{PTUMPM} and VPT~\cite{han2024facing}. Full fine-tuned LLaVA (\ie, LLaVA$_{FT}$~\cite{liu2024visual}) serves as an upper-bound of multimodal evaluation. We report the performance of M$^2$PT under two different settings, with M$^2$PT$_{10/10}$ and M$^2$PT$_{10/20}$.

There are several key observations from these results.
\textbf{First}, M$^2$PT achieves the best performance among all PEFT approaches in most cases, demonstrating the effective design of visual and textual prompts. For example, on MME task, M$^2$PT demonstrates a significant improvement of 6.90\% and 7.51\% compared to two strong prompt tuning methods, Aprompt and VPT, respectively. This highlights the limitation of existing prompt tuning approaches that primarily focus on single modality, failing to capture the cross-modality interactions. In contrast, the interaction layer together with the multimodal LLM employed by our approach successfully bridges this gap, resulting in enhanced performance.
\textbf{Second}, the performance of M$^2$PT reaches \textbf{94.75\%} of the full finetuning performance while considering only \textbf{0.09\%} of the total model parameters, demonstrating the parameter efficiency of M$^2$PT. Moreover, we observe that M$^2$PT outperforms the full finetuning LLaVA on VSR, MNIST and POPE tasks, showing its strong capability in zero/few shot learning. This is consistent with the observations in several previous works~\cite{han20232vpt,YangWWQFCKWXL23}. \textbf{Third}, it can be seen that M$^2$PT does not perform very well on the visual entailment task, SNLI-VE. Our hypothesis is that logical relationships or causal scenario understanding is critical in this task, which might not be fully captured by prompt tuning based approaches.

\section{Analysis and Discussion}\label{subsec:ablation} 
\paragraph{Attention Activation Pattern Analysis.}
Following common practice~\cite{massiveattention}, we extract and discuss the activation maps from the attention block of MLLMs, and investigate the influence of visual and textual prompts in Fig.~\ref{fig:attention}. 
We randomly select two samples from the MME dataset and visualize their corresponding activation maps of the attention block in the last layers of both visual encoder (Fig.~\ref{fig:attention}(a)) and LLM (Fig.~\ref{fig:attention}(b)).
To analyze the impact of textual and visual prompts to the frozen components, we categorize them, according to LLaVa's model structure, into textual prompts, system text tokens, visual prompts, image tokens and instructions. Several findings can be observed. 
\begin{figure}[t]
    \centering
    \includegraphics[width=0.43\textwidth]{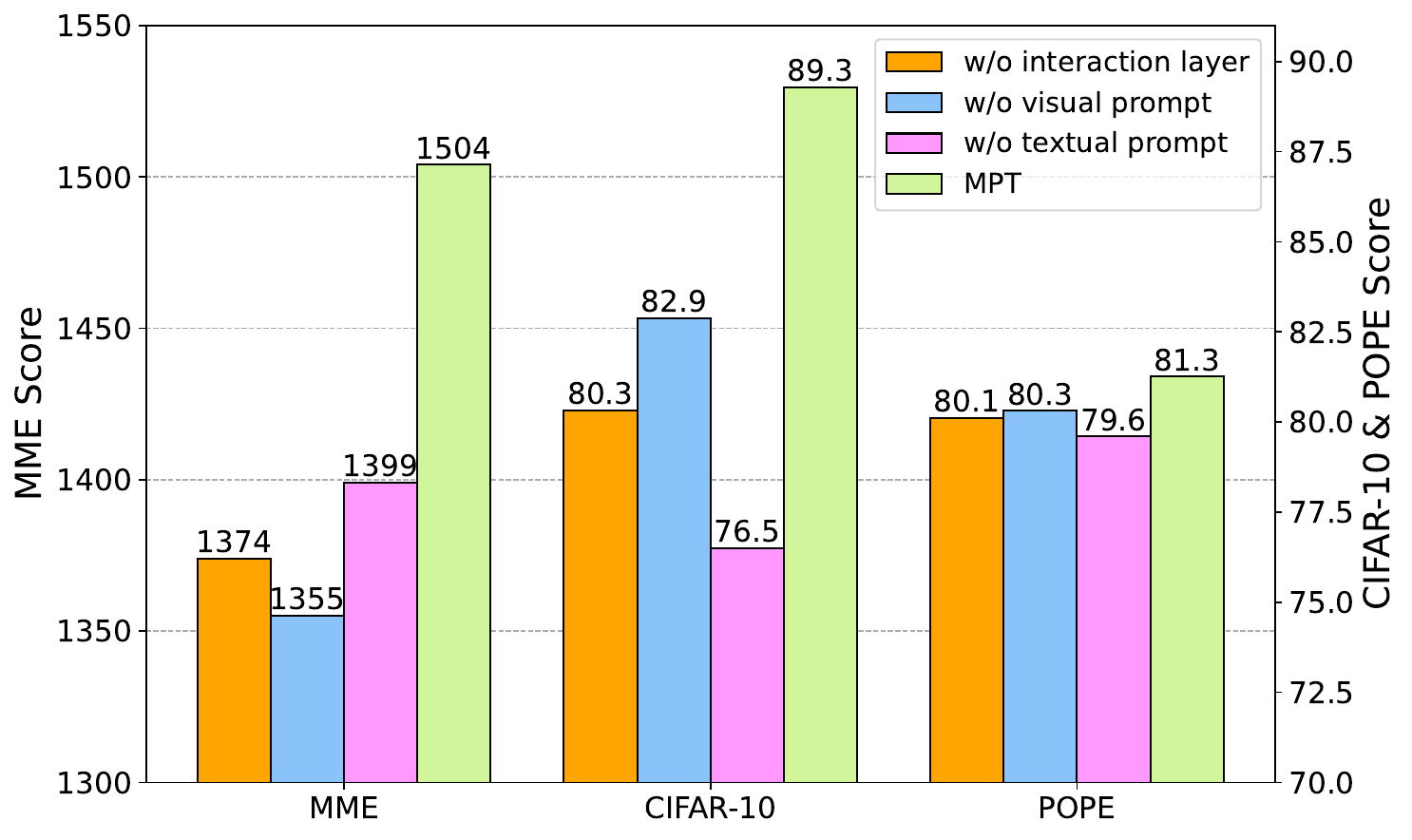}
    \vspace{-3mm}
    \caption{\textbf{Impact of Different Components.} 
    } 
    \vspace{-6mm}
    \label{fig:modality_exp}
\end{figure}
\textbf{First}, in the \emph{LLM} attention activation map, we observe that the token regions corresponding to textual prompts exhibit elevated activation levels, indicating their significant role in shaping the model's responses. 
The activation levels of visual prompts within the LLM, while comparatively lower, remain notable relative to most other regions. This suggests a secondary yet substantial role in multimodal inference.
\textbf{Second}, in the activation maps from \emph{Visual Encoder}, the activation levels associated with visual prompts are noticeably higher than those of most other tokens, underscoring the critical role of visual prompts in processing visual information during tuning. 
These observations support that visual prompts effectively interact with the textual prompts, enhancing the alignment between modalities and thereby improving the model's performance on the zero-shot instruction learning. Our component analysis study below further strengthens this claim quantitatively.

\paragraph{Impact of Different Components.} 
To investigate the impact of different components in M$^2$PT across various datasets (\ie, visual prompts, textual prompts, and interaction layer), we conducted component ablation experiments in Fig.~\ref{fig:modality_exp} by removing each component at a time from M$^2$PT$_{10/20}$.
The results demonstrate that the model performance drops when any of the trainable prompts is
removed, which aligns with our expectations. Moreover, the model performance also decreases without the trainable interaction layer, indicating the importance of this layer. Furthermore, 
we observe that the importance of different components varies across the tasks. For example, textual prompts play the most important role in CIFAR-10 and POPE, while visual prompts lift the performance most on MME. Once again, it is worth noting that combining the multimodal prompts with the interaction layer in M$^2$PT leads to the best performance.

\begin{figure}[t]
    \hspace{-0.7em}
    \includegraphics[width=0.50\textwidth]{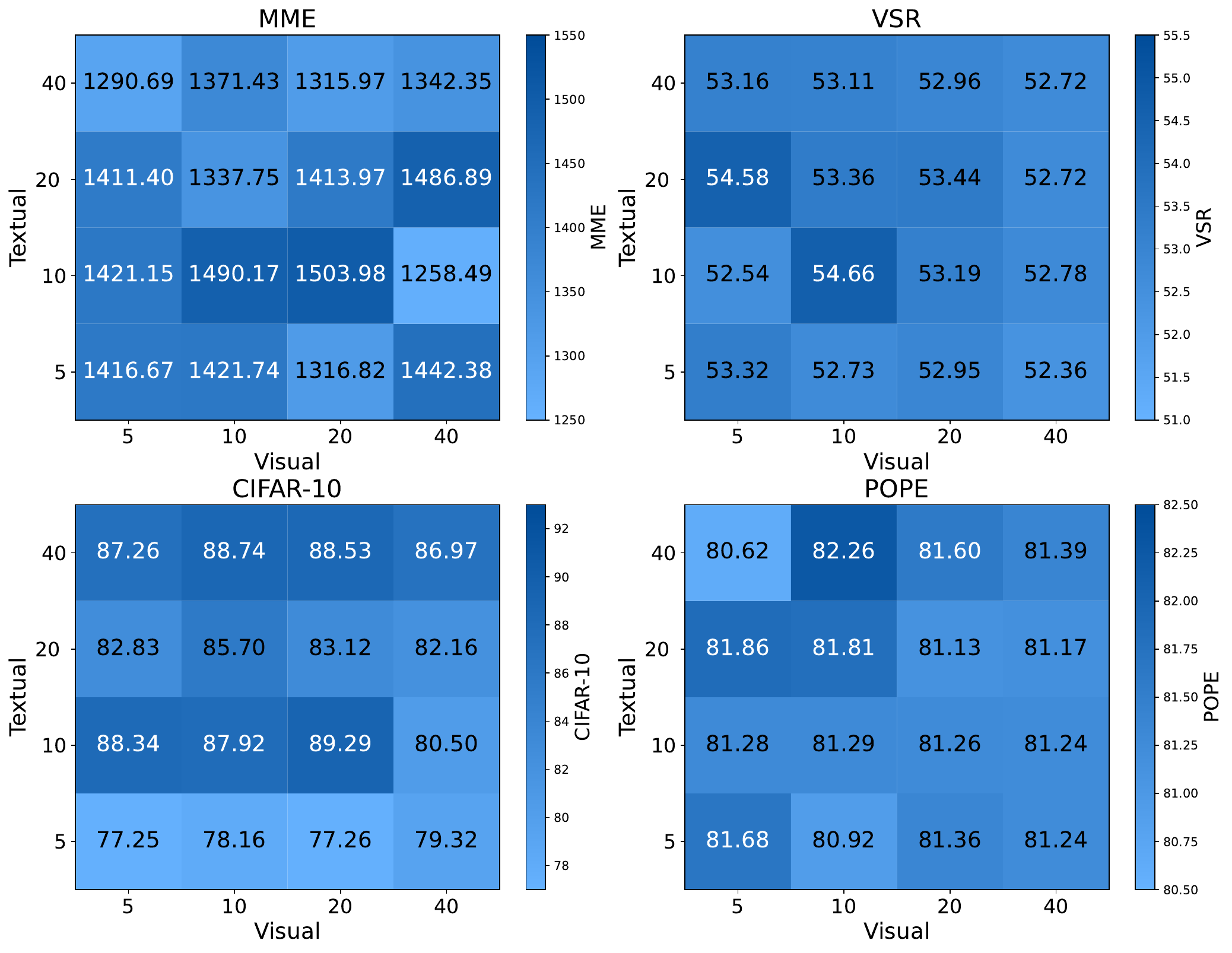}
    \vspace{-6mm}
    \caption{\textbf{Performance of Different Prompt Length.} Each cell in the map corresponds to the score of a model with a textual prompt length (row) and a visual prompt length. A darker hue indicates a higher score, whereas a lighter hue signifies a lower score. 
    }
    \vspace{-5mm}
    \label{fig:prompt_length}
\end{figure}

\paragraph{Impact of Prompt Length.} 
In M$^2$PT, prompt length is the \textit{only} hyperparameter that needs to be tuned. To further analyze the impact of different prompt lengths on model performance, we conduct a comprehensive study on the lengths of visual and textual prompts on the Vision-Flan dataset to better understand their properties. Following common practice~\cite{han20232vpt, jia2022visual, han2024facing}, we use the grid search to span a range from $5$ to $40$ on both visual and textual prompt lengths, reported in Fig.~\ref{fig:prompt_length}. 
We stop at a prompt length of $40$ because performance saturation is observed around this point. Thus, extending the prompt length further would result in increased parameter usage without significant performance gains (\ie, M$^2$PT shows a slight decrease in performance on MME. We argue that this may be due to overparameterization~\cite{han20232vpt,jia2022visual}). When visual prompt length extends from $5$ to $20$, and textual prompt length extends from $5$ to $10$, noticeable performance gains can be observed.
It can be seen that there is no universal optimal prompt length that consistently achieves the
best performance across different tasks. For instance, the optimal performance of the model on MME is achieved with a configuration of $20$ visual prompts and $10$ textual prompts, while $10$ visual prompts and $40$ textual prompts achieve the highest performance on POPE. We hypothesize that different tasks exhibit distinct data distributions, with `difficult' tasks potentially requiring longer prompts to effectively capture the underlying patterns. Nonetheless, we observed that the performance of M$^2$PT remains relatively stable within a certain range of prompt lengths.

\begin{table}[t]
\caption{\textbf{Prompt Location Experiment} (M$^2$PT$_{10/20}$).} 
\vspace{-3mm}
\label{table:prompt-location}
\begin{center}
\begin{small}
\renewcommand\arraystretch{1.15}
\resizebox{0.42\textwidth}{!}{
\begin{tabular}{l|c c c} 
\toprule
\rowcolor{mygray}
Prompt Location &  MME & CIFAT-10 & POPE  \\ 
\midrule
(a) First Layer &  1320.99 & 82.96 & 79.48 \\
(b) Odd Layer   & 1396.87 & \underline{87.65} & 75.79  \\
(c) Top Half   & \underline{1473.42}& 85.82 & \underline{80.31}  \\
(d) Latter Half   & 1249.39  &  83.72 & 79.34  \\
(e) All &  \textbf{1503.98} & \textbf{89.29} & \textbf{81.26} \\
\bottomrule
\end{tabular}}
\end{small}
\end{center}
\vspace{-1.5em}
\end{table}
\paragraph{Impact of Prompt Location.} 
Following~\cite{jia2022visual}, Table~\ref{table:prompt-location} studies the insertion of prompts at different layers.
We design five distinct settings in which prompts are integrated into both the Vision Encoder and LLM but at different locations. Specifically, we introduce prompts into:
(a) the first layer; (b) every odd-number layer (\ie, $[1,3,\dots,23] \in N$, $[1, 3,\dots,31] \in M$); (c) the first half of the layers (\ie, $[$1-12$] \in N$, $[$1-16$] \in M$); (d) the latter half of the layers (\ie, $[$12-24$] \in N$, $[$16-32$] \in M$); and (e) all layers.
Each variant reports the best prompt length combination selected with MME evaluation. Generally, M$^2$PT’s performance is positively correlated with prompt depth. Yet the accuracy drops when inserting prompts from top to bottom, suggesting that prompts at earlier layers matter more than those at latter layers, which is consistent with the observations in~\cite{jia2022visual,aprompt}.

\vspace{-3mm}
\begin{figure}[!tbh]
    \centering
    \includegraphics[width=0.43\textwidth]{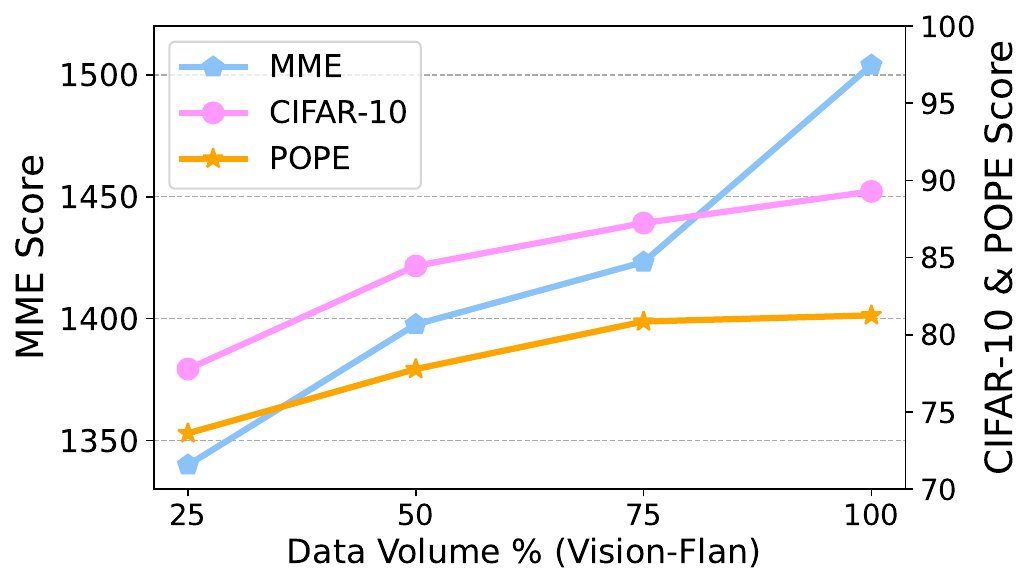}
    \vspace{-3mm}
    \caption{\textbf{Data Volume Experiment} (M$^2$PT$_{10/20}$).}   
    \vspace{-5mm}
    \label{fig:volume_exp}
\end{figure}

\begin{figure}[!tbh]
    \centering
    \includegraphics[width=0.43\textwidth]{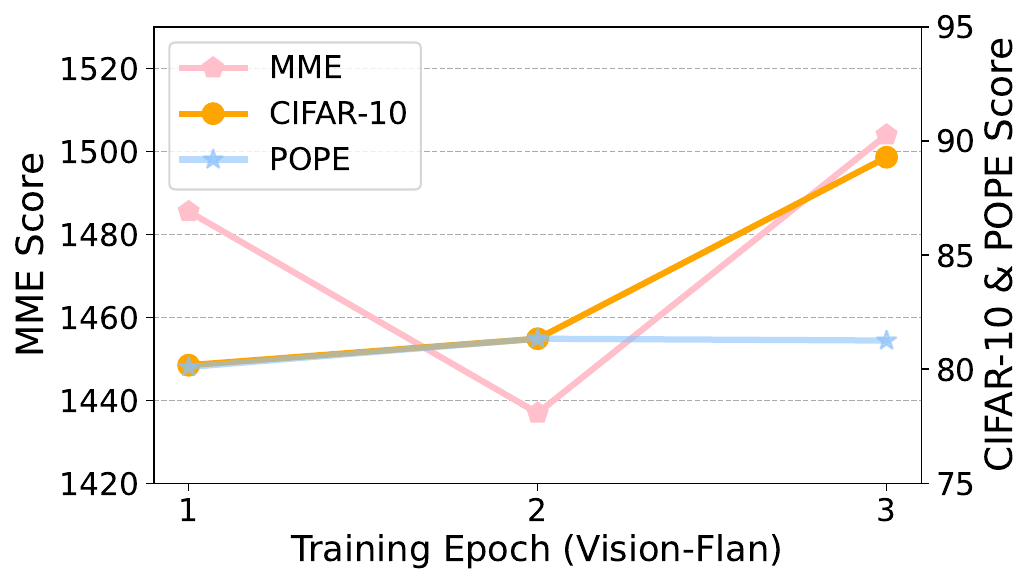}
    \vspace{-3mm}
    \caption{\textbf{Training Epoch Experiment} (M$^2$PT$_{10/20}$).}   
    \vspace{-5mm}
    \label{fig:epoch_exp}
\end{figure}

\paragraph{Effect of Data Volume and Training Epoch.} 
In Fig.~\ref{fig:volume_exp}, we randomly sample data from \textit{Vision-Flan} at different scales (\ie, 25\%, 50\%, 75\%, 100\%) to evaluate the performance of the model with limited data.
The results demonstrate that M$^2$PT maintains excellent performance despite significant data reduction, highlighting efficiency and robustness regarding data quantity. This property indicates that M$^2$PT exhibits substantial tolerance to data scale, suggesting a promising future for real-world applications with constrained data.

In Fig.~\ref{fig:epoch_exp}, we analyze the M$^2$PT's performance on different training epochs. This experiment is conducted using the optimal hyperparameter combination of $10$ textual and $20$ visual prompts. It can be seen that the model performance generally improves with more training epochs, confirming that M$^2$PT can achieve remarkable performance after sufficient training and demonstrate robustness in adapting to different training epochs. 

\vspace{-3mm}

\begin{figure}[t]
    \centering
    \includegraphics[width=0.45\textwidth]{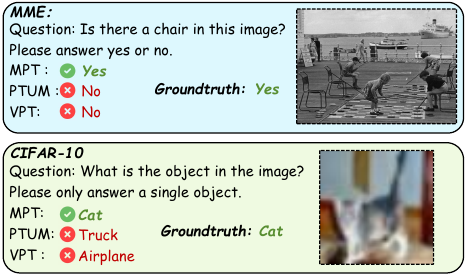}
    \vspace{-5pt}
    \caption{\textbf{Winning cases} on \textit{MME} and \textit{CIFAR-10}.}   
    \vspace{-2mm}
    \label{fig:case_study2}
\end{figure}
\paragraph{Case Study.} 
In Fig.~\ref{fig:case_study2}, we present cases where M$^2$PT demonstrates success on the \textit{MME} and \textit{CIFAR-10}, while other approaches fail (\ie, VPT, PTUM). For example, on MME, while M$^2$PT correctly identifies ``a chair'' in the image, both VPT and PTUM fail to capture this concept and respond with the wrong answer. On \textit{CIFAR-10}, VPT and PTUM both misidentify ``cat'' as a different category.
We posit that the insufficiency of VPT may stem from its inadequate understanding of logical relationships or causal scenarios. PTUM employs single-modality tuning exclusively, rendering it inadequate for managing complex multi-modal behavior and capturing interactions between modalities. In sharp contrast, M$^2$PT takes an aspect of multimodal, enhancing both visual modality comprehension and textual modality causal inference.

\begin{figure}[t]
    \centering
    \includegraphics[width=0.45\textwidth]{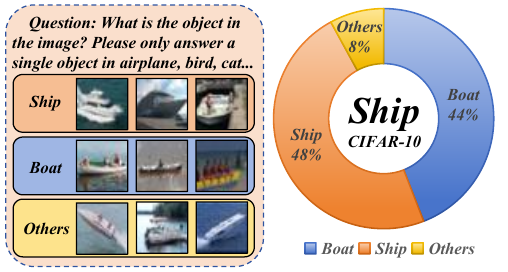}
    \vspace{-2mm}
    \caption{\textbf{Failure cases} study on \textit{CIFAR-10}.}   
    \vspace{-3mm}
    \label{fig:case_study}
\end{figure}

We further conduct failure case studies for M$^2$PT on CIFAR-10 dataset. The results in Fig.~\ref{fig:case_study} indicate that the predominant misclassification within the CIFAR-10 (\ie, 44\% or 443 examples of all such misclassification) is the misidentification of ``ships'' as ``boats''. This high error rate could potentially be attributed to the limited resolution of the images within the CIFAR-10, coupled with the inherent similarity in the semantic characteristics between ``ship'' and ``boat''. Additional case studies and discussions are provided in Appendix \ref{Appendix:case-study}.

\section{Conclusion}\label{sec:conclusion}
We introduce M$^2$PT, a novel framework in multimodal prompt tuning for zero-shot instruction learning. Our framework offers several advantages: \textbf{i)} it introduces visual and textual prompts elegantly into vision encoder and LLM, respectively, enabling fast and accurate multimodal adaptation; \textbf{ii)} it synergizes modalities by cross-modality interaction, enjoying coherent integration from multimodal perspectives; and \textbf{iii)} it significantly reduces the number of trainable parameters compared to conventional finetuning methods, while maintaining robust performance across zero-shot tasks. As a whole, we conclude that the outcomes elucidated in this paper impart essential understandings and necessitate further exploration within this realm.

\section*{Limitations}
For potential limitations, M$^2$PT requires two hyperparameters on prompt length searching (\ie, Visual Prompt, Textual Prompt). Though in practice, we find both lengths vary into a relatively narrow range (see \S\ref{subsec:ablation}), and are sufficient enough to outperform $all$ current methods, there is still possible integration~\cite{DBLP:conf/icml/HeZTGDAZLCMCC22} of a local searching network to generate optimal combinations of lengths. 
Another potential limitation is that M$^2$PT, akin to other PEFT approaches~\cite{han20232vpt, jia2022visual}, lacks \textit{ad-hoc} explainability~\cite{biehl2016prototype, wang2022visual}. While in \S\ref{sec:exp}, we demonstrates that activation maps from MLLMs are significantly influenced by visual and textual prompts, further research is necessary to elucidate the underlying nature of these prompts.

\section{Acknowledgements}
This research was supported by the National Science Foundation under Grant
No. 2242243.
The views and conclusions contained herein are those of the authors and should not be interpreted as necessarily representing the official policies or endorsements, either expressed or implied, of U.S. Naval Research Laboratory (NRL) or the U.S. Government. The U.S. Government is authorized to reproduce and distribute reprints for Government purposes notwithstanding any copyright notation herein.

\bibliography{main}

\clearpage

\appendix
\renewcommand{\thesection}{S\arabic{section}}
\renewcommand{\thetable}{S\arabic{table}}
\renewcommand{\thefigure}{S\arabic{figure}}
\setcounter{table}{0}
\setcounter{figure}{0}
\centerline{\textbf{SUMMARY OF THE APPENDIX}} 
\vspace{0.5em}

This appendix contains additional experimental results and discussions of our work, organized as:
\begin{itemize}[leftmargin=*]
\setlength{\itemsep}{0pt}
  \item \S\ref{appendix:data-detailed} includes additional introduction on datasets applied in our paper.
  \item \S\ref{Appendix:Training-detail} provides more training details on our proposed M$^2$PT.
  \item \S\ref{Appendix:Evaluation-Metrix} presents more details on the evaluation metrics applied in \S\ref{sec:exp}.
  \item \S\ref{Appendix:case-study} further includes more case studies of M$^2$PT on its perceptual proficiency.
  \item \S\ref{Appendix:connection} discuss the relations of M$^2$PT with previous works and their connections.
  \item \S\ref{Appendix:init} provides discussions and additional results on the impact of prompt initialization.
  \item \S\ref{Appendix:Discussion} provides discussions on licenses, reproducibility, social impact, and directions of our future work.
\end{itemize}
\section{Data Statistics}\label{appendix:data-detailed}
Table \ref{appendix:data-detailed} shows details of 9 multimodal datasets for our finetuning and evaluation. Vision-Flan~\cite{xu2024vision} includes 191 different multimodal tasks which is ideal for our finetuning process. Each multimodal tasks contains up to 1,000 instances, resulting in a total of 191,105 instances. MME~\cite{yin2023survey} serves as our comprehensive multimodal evaluation benchmark, measuring both perception and cognition capabilities across 14 subtasks. We further leverage 7 multimodal datasets for our evaluation.
Specifically, for Optical Character Recognition, we utilize the \textit{Text-VQA}~\cite{singh2019towards}, and for reasoning, we employ the Visual Spatial Reasoning (\textit{VSR})~\cite{liu2023visual}. Following~\cite{zhai2023investigating, shen2024multimodal}, the perception capability is tested on \textit{CIFAR-10/100}~\cite{krizhevsky2009learning} and \textit{MNIST}~\cite{deng2012mnist}. \textit{SNLI-VE}~\cite{xie2019visual} evaluates Visual Entailment capabilities, while the \textit{POPE}~\cite{li2023evaluating} dataset examines the tendency towards object hallucination. The MME metric is the sum of accuracy values across all subtasks, while for the other 7 multimodal evaluation datasets, the metric used is just accuracy.

\begin{table}[!ht]
\caption{\textbf{Multimodal Dataset Details}}
\label{table:data_detail}
\begin{adjustbox}{width=0.9\width,center}
\begin{tabular}{c c c} 
\toprule
\rowcolor{mygray}
Dataset & Examples & Task Categories \\
\midrule
Vision-Flan & 191K & Diverse \\
MME & 2374 & Diverse \\
Text-VQA & 5000 & OCR\\
VSR & 1222  & Reasoning   \\
SNLI-VE & 17901 & Entailment  \\
CIFAR-10 & 10000  & Perception \\
CIFAR-100 & 10000  & Perception \\
MNIST & 10000  & Perception \\
POPE & 9000  & Object Hallucination\\
\bottomrule
\end{tabular}
\end{adjustbox}
\end{table}

\section{Implementation Details}\label{Appendix:Training-detail}
\label{appendix:training}
Stage-one LLaVA~\cite{liu2024visual} is utilized as our pre-trained multimodal model. Specifically, we employ LLaVA with Vicuna-7B-v1.3 as the base LLM and CLIP-L as the vision encoder for all variants. The finetuning process for M$^2$PT$_{10/20}$ takes approximately 9 hours on 4 A100 GPUs (40G), with a batch size of 8 per GPU and a gradient accumulation step of 4 (128 in total), with the same data preprocess and normalize method as LLaVA. Additional configurations of M$^2$PT are shown in Table \ref{Appendix:Training-detail}. For LoRA, we directly import the best results from \cite{shen2024multimodal}. For the prompt tuning baselines, APrompt~\cite{aprompt} and PTUM~\cite{PTUMPM} add textual/attention prompts into the LLM, while VPT~\cite{han2024facing} only appends visual prompts to the vision encoder. We use the optimal settings in the original papers to train their models, with grid search on the best learning rate. For the other configuration, we adopt LLaVA’s~\cite{liu2024visual} default settings as provided in its codebase.

\begin{table}[!ht]
\caption{\textbf{Hyperparameters and configurations.} }
\label{table:training_details}
\begin{center}
\begin{small}
\resizebox{0.32\textwidth}{!}{
\begin{tabular}{l c} 
\toprule
Learning Rate & $7e^{-4}$ \\
Batch Size & 128 \\
Lr Scheduler &  cosine  \\
Warmup Ratio & 0.03   \\
Activation Type & bfloat16  \\
Weight Decay & 0  \\
Model Max Length & 1024 \\
\bottomrule
\end{tabular}}
\end{small}
\end{center}
\vspace{-1.5em}
\end{table}
\section{Evaluation Metrics}\label{Appendix:Evaluation-Metrix}
\label{appendix:eval}

For evluation, we utilize MME~\cite{yin2023survey} and the other 7 multimodal datasets (see \S\ref{appendix:data-detailed}). For MME, we employ the official evaluation tool~\cite{yin2023survey} of MME, including the Perception and Cognition metrics. For the other 7 multimodal datasets, following~\cite{shen2024multimodal}, we employ the same prompt template to guide Vicuna-13B-v1.5~\cite{zheng2024judging} in evaluating the accuracy of each prediction, considering the specified task instructions and the groundtruth target output. All tasks are classification tasks and we calculated the final score of each multimodal dataset based on the percentage of vicuna 13b answering ``Yes.''

\section{More Case Study}\label{Appendix:case-study}
\begin{figure*}[!tbh]
    \centering
    \includegraphics[width=1\textwidth]{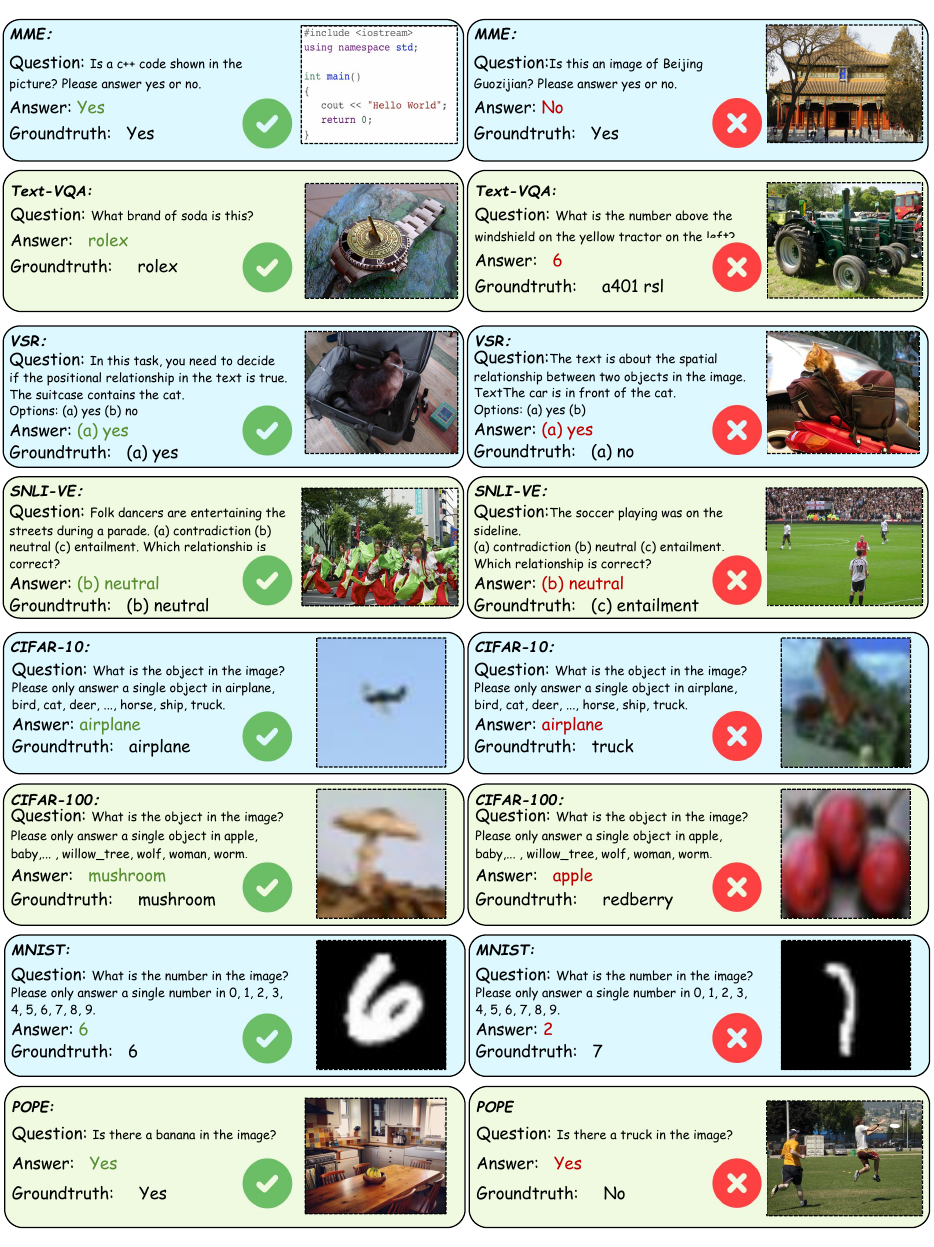}
    \caption{More Case study on 8 \textit{Zero-Shot} datasets~(M$^2$PT$_{10/20}$).}   
    \label{fig:case_study3}
\end{figure*}

To further investigate the model's performance and delineate instances of its suboptimal functioning, we conduct an in-depth visual assessment of a sample cohort drawn from eight distinct \textit{Zero-Shot} datasets, as illustrated in Fig.~\ref{fig:case_study3}. This visualization facilitates a comprehensive understanding of the model's efficacy across a diverse array of tasks and data, while concurrently revealing potential constraints inherent to the model and the underlying causes of its occasional shortcomings. From the listed failure cases, we summarize 2 failure patterns, which are: (a) Small objects perception failure in \textit{Text-VQA}, \textit{CIFAR-10}, \textit{MNIST}. Small targets in images pose a challenge to perception, impacting the quality of VQA. Further research is essential to enhance accuracy in diverse contexts. (b) Semantic similar failure in \textit{CIFAR-10}, \textit{MNIST}, \textit{POPE}, the inability to distinguish between semantically similar objects results in MLLMs generating wrong answers. Developing methods that can effectively differentiate between similar objects is essential for real-world applications. This involves enhancing the model's capacity to learn fine-grained features and contextual information, thereby improving its overall accuracy and robustness.

\section{Discussion with Previous Works}\label{Appendix:connection}
This section provides discussion that connects M$^2$PT with previous methods. 
If we remove the textual prompts and the interaction layer, our model architecture degenerates to the visual prompt tuning approaches~\cite{jia2022visual,han2024facing, han20232vpt}. If we completely freeze the vision encoder by only introducing the textual prompts, our model is similar to those traditional prompt tuning methods~\cite{lester2021power,ma2022xprompt,PTUMPM} in LLM. Moreover, if we further incorporate the attention prompts in the LLM, our model is close to the APrompt approach~\cite{aprompt,han20232vpt}. Nevertheless, very limited work focuses on the efficient tuning of multimodal large language models.

\begin{table}[!ht]
\caption{\textbf{Initialization Comparison.} We compare the performance of M$^2$PT under different initialization methods in the setting of best prompt combination~(\ie, 10 and 20)}

\label{table:initialization}
\begin{adjustbox}{width=0.9\width,center}
\begin{tabular}{c c c} 
\toprule
\rowcolor{mygray}
Method & Initialization & MME \\
\midrule
M$^2$PT$_{10/20}$ & Random & 1405.67 \\
M$^2$PT$_{10/20}$ & Xavier & 1503.98 \\
\bottomrule
\end{tabular}
\end{adjustbox}
\end{table}

\section{Prompt Initialization}\label{Appendix:init}
Table~\ref{table:initialization} reports the performance of M$^2$PT with respect to x widely adopted initialization methods: 
\textit{Xavier}~\cite{xavier} and random on \textit{MME}. The results show that \textit{Xavier} generally provides more stable and preferable performances. In conclusion, M$^2$PT shows robustness on different initialization methods and is able to achieve comparable performance with full finetuning.

\section{Discussion}
\label{Appendix:Discussion}

\subsection{Asset License and Consent}
\label{Appendix:License}

The majority of VPT~\cite{jia2022visual}, \href{https://textvqa.org/dataset/}{\textit{Text-VQA}},  is licensed under \href{https://github.com/KMnP/vpt/blob/main/LICENSE}{CC-BY-NC 4.0}. Portions of \cite{jia2022visual} are available under separate license terms: \href{https://github.com/google-research/task_adaptation}{google-research/task\_adaptation}, \href{https://github.com/huggingface/transformers}{huggingface/transformers}, \href{https://github.com/haotian-liu/LLaVA}{LLaVA},  \href{https://github.com/meta-llama/llama}{Vicuna}, \href{https://github.com/cambridgeltl/visual-spatial-reasoning/blob/master/LICENSE}{\textit{VSR}} is licensed under \href{https://www.apache.org/licenses/LICENSE-2.0}{Apache-2.0}. \href{https://github.com/jeonsworld/ViT-pytorch}{ViT-pytorch}~\cite{dosovitskiy2020image} and \href{https://github.com/RUCAIBox/POPE?tab=MIT-1-ov-file}{POPE} is licensed under \href{https://opensource.org/license/mit/}{MIT}; \href{https://github.com/necla-ml/SNLI-VE/blob/master/LICENSE}{SNLI-VE} are under \href{https://opensource.org/license/bsd-3-clause}{BSD 3-Clause}

\subsection{Reproducibility}
\label{appendix:reproduce}
M$^2$PT is implemented in Pytorch~\cite{NEURIPS2019_9015}. Experiments are conducted on NVIDIA A100 GPUs. To guarantee reproducibility, our full implementation shall be publicly released upon paper acceptance.

\subsection{Social Impact}
\label{Appendix:social_impact}
This work introduces M$^2$PT possessing strong performance gains over state-of-the-art baselines in \S\ref{sec:exp}, with considerably low parameter usage for MLLMs. Our approach advances model accuracy, and is valuable in parameter-sensitive training applications, \eg, MLLMs on devices and fast adaptation with limited resources.

\subsection{Potential Risks}
Consider the tuning process of LLM, which has potential risks for energy usage. Finetuning requires significant computational power, leading to high energy use and increased environmental impact. 
\subsection{Future Work}
\label{Appendix:future_work}
Despite M$^2$PT's systemic effectiveness and efficacy during instruction tuning, it also comes with new challenges and unveils some intriguing questions. For instance, incorporating an advanced network into M$^2$PT to search the optimal combinations of prompt lengths (\ie, Visual Prompt, Textual Prompt) might significantly reduce the search space of lengths and lead to further performance gains. Another essential future direction is the design and analysis of network interpretability~\cite{arrieta2020explainable,laugel2019dangers,rudin2019stop} and \textit{ad-hoc} explainability~\cite{biehl2016prototype, wang2022visual}, which limits current adoption of M$^2$PT in decision-critical, real-world applications. Overall, we believe the results presented in this paper warrant further exploration.

\end{document}